\documentclass{OAGM}
\OAGMarXiv{1304.1876}
\usepackage{subfigure}
\usepackage{graphicx}

\title{Pulmonary Vascular Tree Segmentation from Contrast-Enhanced CT Images}

\author{M. Helmberger\textsuperscript{1,2} \and 
		M. Urschler\textsuperscript{1,3} \and 
		M. Pienn\textsuperscript{2} \and 
		Z. B\'{a}lint\textsuperscript{2} \and 
		A. Olschewski\textsuperscript{2,4} \and 
		H. Bischof\textsuperscript{1} \and \\
  \textsuperscript{1}Institute for Computer Graphics and Vision, Graz University of Technology, Austria\\
  \textsuperscript{2}Ludwig Boltzmann Institute for Lung Vascular Research, Graz, Austria\\
  \textsuperscript{3}Ludwig Boltzmann Institute for Clinical-Forensic Imaging, Graz, Austria\\
  \textsuperscript{4}Experimental Anesthesiology, Department of Anesthesia and Intensive Care Medicine,\\
  Medical University of Graz, Austria}

\begin{document}
\maketitle

\begin{abstract}
We present a pulmonary vessel segmentation algorithm, which is fast, fully automatic and robust. It uses a coarse segmentation of the airway tree and a left and right lung labeled volume to restrict a vessel enhancement filter, based on an offset medialness function, to the lungs. We show the application of our algorithm on contrast-enhanced CT images, where we derive a clinical parameter to detect pulmonary hypertension (PH) in patients. Results on a dataset of 24 patients show that quantitative indices derived from the segmentation are applicable to distinguish patients with and without PH. Further work-in-progress results are shown on the VESSEL12 challenge dataset\footnote{http://vessel12.grand-challenge.org/}, which is composed of non-contrast-enhanced scans, where we range in the midfield of participating contestants.
\end{abstract}

\section{Introduction}

Since its introduction in the 1970s, computed tomography (CT) has become an important tool in medical imaging. It is the gold standard in the diagnosis of a large number of different disease entities~\cite{Day01112011}, and further technological progress has strengthened its diagnostic impact leading to an essential role in clinical practice. To gain full benefit of the increasing resolution of CT images, automatic methods are needed to separate important information from diagnostically irrelevant ones.

Automatic segmentation and analysis of the vessels inside the lung (pulmonary vessels) from CT images is widely used for computer aided diagnosis of vascular diseases~\cite{Shikata:2009:SPV:1540564.1807590}, non-rigid image registration~\cite{Maeda2012}, and detection of pulmonary embolism~\cite{Zhou2007}. Our clinical focus is on the detection of pulmonary hypertension (PH), which is a chronic disorder of the pulmonary circulation, marked
by an elevated vascular resistance and elevated mean pulmonary artery
pressure (mPAP)~\cite{Galie2009}. Unlike the systematic circulation, the blood pressure in the pulmonary vessels is very difficult to measure. In order to determine pulmonary pressure, an invasive right heart catheterisation must be performed~\cite{Galie2009}. By the time of diagnosis, PH has usually progressed to late stage and is not reversible any more. By finding a non-invasive way of measuring the pulmonary blood pressure, the number of patients awaiting treatment could be significantly decreased. Therefore, an important aim of clinical PH research is the early diagnosis of pulmonary hypertension.

\begin{figure}[htb]
\centering
\includegraphics[width=0.7\textwidth]{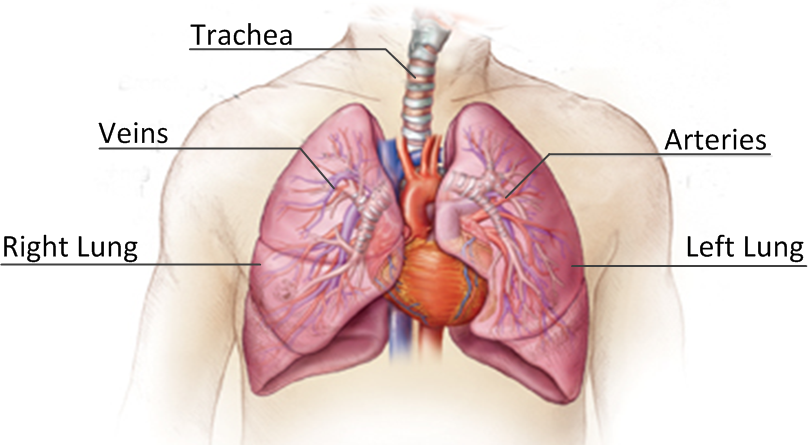}

\caption{Anatomy of the human lung showing the airways, starting with the trachea, left and right lung, the pulmonary arteries in red and the pulmonary veins in blue \footnotesize{(\textcopyright\space www.somersetmedicalcenter.com)}. }
\label{fig:human-lung}
\end{figure}

For determining a measure for pulmonary blood pressure, a segmentation of the blood vessels inside the lung is needed. We present a robust algorithm that combines lung- and airway-segmentation, together with a sophisticated vessel enhancement filter to obtain a proper segmentation of the left and right pulmonary vessel trees, even in patients showing severe pathologies. A schematic of the anatomy of the human lung is shown in Figure~\ref{fig:human-lung}.

The algorithm is fully automatic, computationally efficient and able to handle large datasets. It was tested on phantom data, the publicly available VESSEL12
challenge dataset, and CT data from 24 patients from a clinical PH study.

\subsection{Related Work}

Several 3D vessel segmentation algorithms have been presented in the literature up till now. An overview can be found in~\cite{LesageABF09}. Typical approaches include threshold-based algorithms \cite{AtillaP.Kiraly2004} or fuzzy methods \cite{Kaftan2008}. These approaches have in common that an intensity model is utilized to detect the vessels. Frangi et al.~\cite{Frangi:1998} presented a technique based on the eigenvalue analysis of the Hessian matrix. This approach was later refined in the popular approaches of~\cite{Shikata:2009:SPV:1540564.1807590} and~\cite{Krissian2000130}, who also take the eigenvectors of the Hessian matrix into account. In \cite{Agam2005} they apply vessel-enhancement, junction-enhancement and nodule-enhancement filters, based on an eigenvector analysis from the gradient vector field. Most of these approaches work well on controls, however, for patients showing pathologies, robust extraction of vascular structures is still an open issue, especially in the case of pulmonary hypertension where vessel pruning occurs~\cite{TJ2011}.

\section{Method}

\begin{figure}[htb]
\centering
\includegraphics[width=\textwidth]{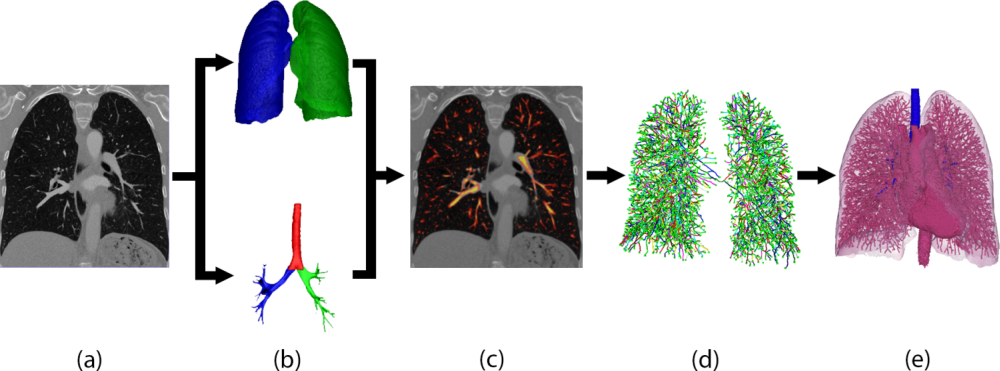}

\caption{Flowchart of the vessel segmentation algorithm. (a) input CT-image, (b) lung- and airway segmentation, (c) vessel enhancement filter restricted to the lung (yellow: high vessel probability), (d) centerline reconnection, (e) vessel segmentation}
\label{fig:Flowchart-of-the_vessel_segmentation_algorithm}
\end{figure}

At the core of our method is a multi-scale vessel enhancement (VE)
filter based on the Hessian matrix. It is similar 
to~\cite{Pock05multiscalemedialness} in using the eigenvectors of the Hessian matrix to detect candidate voxels inside the vessels, and computing an offset-medialness boundary measure perpendicular to the estimated vessel direction~\cite{Krissian2000130}. The VE response (i.e. medialness) is limited to the right and left lung, which is derived from an intensity-based lung segmentation and a coarse airway tree segmentation with a labelling of left and right main bronchi. After centerline detection from the VE response, a coarse radius estimation of the vessel is obtained using a spherical ray-casting approach. The final segmentation is the output of a globally optimal geodesic active contour model based on total
variation and a data term derived from the medialness~\cite{Reinbacher2010}.
Figure~\ref{fig:Flowchart-of-the_vessel_segmentation_algorithm} shows
the flowchart of our automatic vessel segmentation approach. 

\subsection{Lung and airway segmentation \label{sub:lung_airway_seg}}

A prerequisite for our vascular tree extraction is a segmentation of left and right lungs, reactively, to restrict the reconnection of the vessel centerlines. 
A coarse airway segmentation and labelling of the main bronchi initiates this process. The labeled airway tree is subsequently used to label a coarse, threshold based, lung segmentation. Segmenting the airways also helps in removing false positives of the vessel enhancement filter (see Section~\ref{sub:Vessel-enhancement}), since the intensity contrast of the airway border and blood vessels is very low, thus leading to incorrect detection of blood vessels at the airway walls.

\begin{figure}[htb]
	\centering
	\subfigure[]{
		\includegraphics[height=3.2cm]{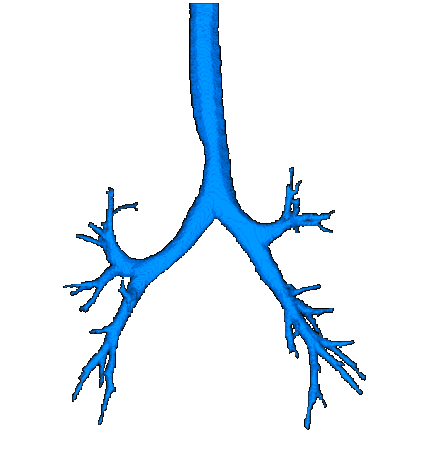}
		}
		\hspace*{1cm}
	\subfigure[]{
		\includegraphics[height=3.2cm]{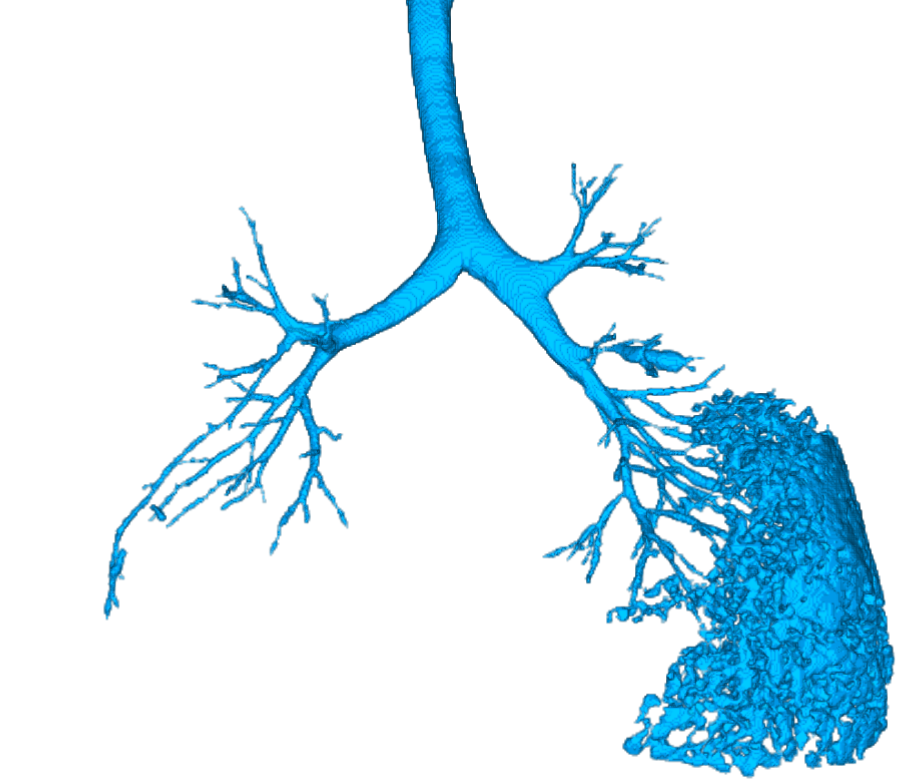}
		}
	\caption{(a) representative result of the automatic airway segmentation from an example patient, (b) leaked airway segmentation}
	\label{fig:Results-of-theairway}
\end{figure}

We automatically detect the airway on the top-most slice of the contrast-enhanced volume, which is a dark circle surrounded by high-intensity tissue, to get a seed point for an iterative 3D region growing algorithm. For the region growing
two thresholds are defined: $th_{min}=I(\mathbf{x_{s}})-1\,\textrm{HU}$
and $th_{max}=I(\mathbf{x_{s}})+1\,\textrm{HU}$, where $I$
is the CT image, $\mathbf{x_{s}}$ is the seed point and HU denotes a Hounsfield Unit. All $N_{6}$
connected voxels which fulfill $th_{min}<I(\mathbf{x})<th_{max}$
are added to the segmentation. Then the thresholds are updated ($th_{min}=th{}_{min}-1\,\textrm{HU}$
and $th_{max}=th_{max}+1\,\textrm{HU}$) and region growing is restarted
with the new segmentation as seed. Convergence of this iterative procedure combines two stopping criteria. We check whether the number of \emph{total voxels}, or the number of  \emph{edge voxels} of the current segmentation is three times larger than the average of all of the previous voxel counts. In this case a leakage is detected (see Figure~\ref{fig:Results-of-theairway}b), and we produce the final airway tree segmentation with the restricted threshold range from the previous iteration.  Figures~\ref{fig:Results-of-theairway}a shows an example of airway segmentation obtained from one patient in the clinical PH study.

After coarse airway segmentation, we perform a left and right lung segmentation to identify a region of interest for later vessel detection. A coarse lung segmentation is obtained using Otsu's optimal thresholding method~\cite{otsu_ieeetsmc_1979}. With a
connected component analysis, the lung is selected and a 3D hole filling is applied to include vascular structures. The two
lungs always merge through the airways, but in some datasets the border
between right and left lung is hardly visible, resulting in connected
lungs. The airway segmentation is used to separate the coarse lung segmentation. Using a graph representation of the skeleton from the airway segmentation, we detect the carina (where the trachea splits into the left and right main bronchi), and assign different labels to the trachea, right and left bronchi. To label the voxels in the coarse lung segmentation, we calculate shortest paths to the labeled airway tree, thus, splitting it into left and right lung. As a cost function $I_{c}$ for the shortest path algorithm, we use the gradient
magnitude of the CT image $|\nabla I|$ and the coarse lung segmentation without the airways, where we give a larger weight to the gradients to prevent connecting via short cuts from touching left and right lungs. We found a fraction of $\frac{4}{5}$ for the gradient weight, and a weight of $\frac{1}{5}$ for the lung segmentation without the airways to perform well in the labeling of our datasets.

\begin{figure}[htb]
	\centering
	
	\subfigure[]{
		\includegraphics[width=0.3\textwidth]{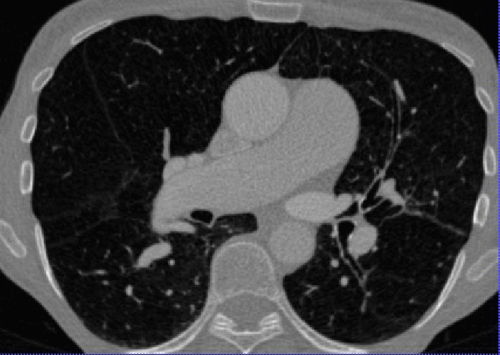}
	}
	\subfigure[]{
		\includegraphics[width=0.3\textwidth]{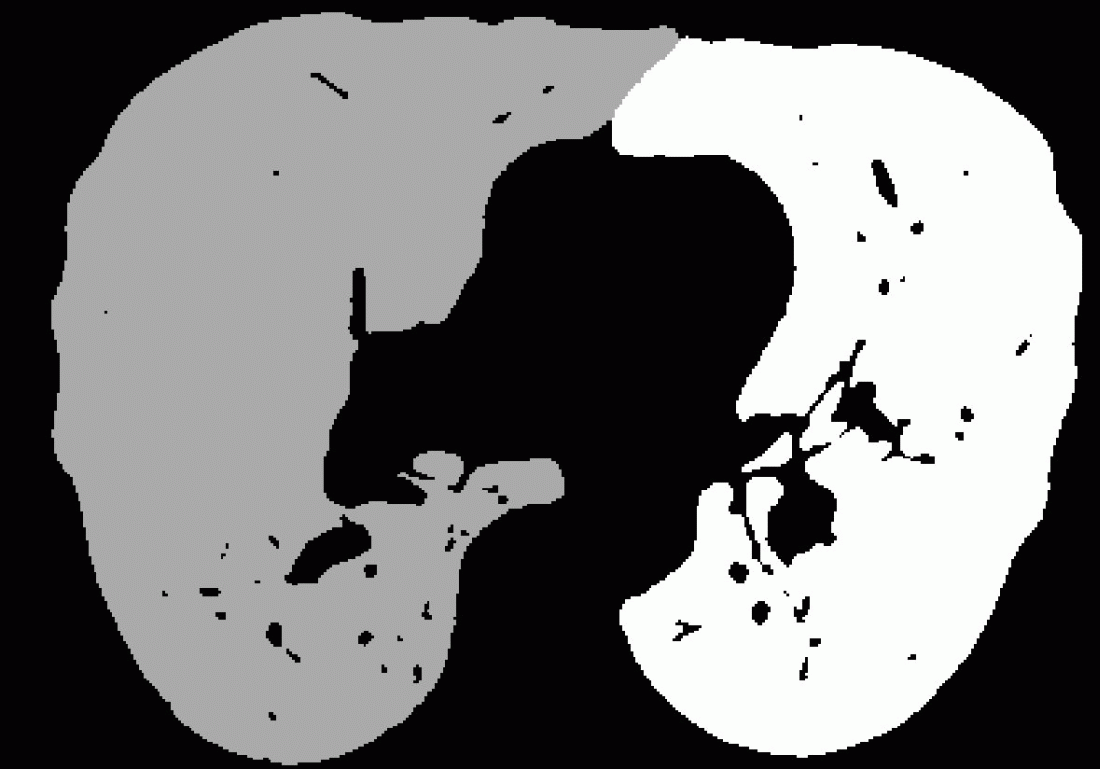}
	}
	\subfigure[]{
		\includegraphics[width=0.3\textwidth]{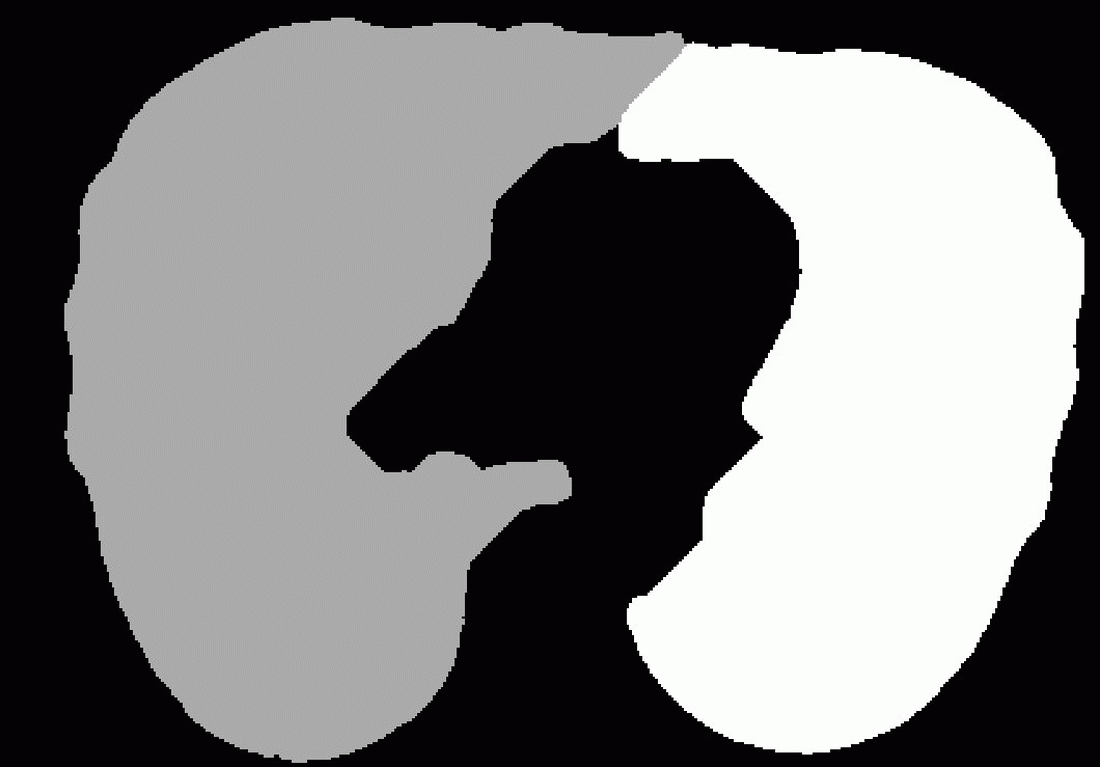}
	}			

	\caption{(a) one 2D slice of an example CT image, (b) coarse lung segmentation after separation, (c) refined lung segmentation, separate left (white) and right (grey) lung}
	\label{fig:LungSeg}
\end{figure}

As a final step, to remove holes caused by vessels and other high
intensity structures inside the lung, the airways are removed from
the lung segmentation, and a morphological closing operation is applied
several times at each lung separately. We use a six-neighbourhood star-shaped structuring element and 10 closing operations. These parameters remain constant for all datasets. This ensures that the lung segmentations contain the whole lung. The different steps of the
lung segmentation can be seen in Figure~\ref{fig:LungSeg}.

\subsection{Vessel enhancement\label{sub:Vessel-enhancement}}

We enhance all vessel-like structures using a modified version of
the vessel enhancement (VE) filter proposed by Pock et al. \cite{Pock05multiscalemedialness}.
We extend the VE filter by using multiple radii in addition to a
multi-scale pyramid with a down-sampling factor of 1.7. With these parameters, the experiments showed improved discrimination between different vessel
radii compared to the widespread use of Gaussian pyramids with a down-sampling
factor of two. Further, to be more robust against noise and to get rotation invariant
derivatives, we compute our gradients for the boundary measure and for the Hessian matrix
using the filter kernels of \cite{Farid04differentiationof}. The airway-
and lung-segmentations from Section~\ref{sub:lung_airway_seg} are used to restrict the vessel
enhancement output to the left and right lungs without the airways, respectively. 

To get the vessel enhancement filter response, we calculate the eigenvalues
$e_{1}$, $e_{2}$, and $e_{3}$ (ordered such that $|e_{1}|\geq|e_{2}|\geq|e_{3}|$) and the associated eigenvectors $\mathbf{v_1}$,
$\mathbf{v_2}$ and $\mathbf{v_3}$ of the Hessian matrix $\mathcal{H^{\sigma}}(\mathbf{x})$
at each scale $\sigma$. To sort out for bright tubular structures
on dark background, we check that $e_1<0$ and $e_2<0$ holds. In points
that fulfill this condition, the smallest eigenvector $\mathbf{v_3}$
gives an estimation for the vessel direction. Perpendicular to the
vessel direction, in the cross section plane of the tube given by
the eigenvectors $\mathbf{v_1}$ and $\mathbf{v_2}$, we evaluate boundary
information along circles of different radii $r$. We define the boundary
gradient $\mathbf{B}(\mathbf{x},\sigma)=\sigma\nabla I^{\sigma}(\mathbf{x})$,
with $I^{\sigma}(\mathbf{x})$ being the CT image convolved with a
Gaussian kernel with variance $\sigma$. An initial response is given
by the median of the boundary contributions $b_{i,\sigma}=|\mathbf{B}(\mathbf{x}+r\mathbf{v}_{\alpha_{i}},\sigma)\mathbf{v}_{\alpha_{i}}|$, with  $i=1\dots\left\lfloor 2\pi r+1\right\rfloor$, which we denote as $R_{0}^{+}$. A problem
of $R_{0}^{+}(\mathbf{x},r,\sigma)$ is, that it also produces responses at isolated
edges. To avoid this, a measure of symmetry is introduced:

\[
S(\mathbf{x},r)=1-\frac{s(\mathbf{x},r)}{\overline{b}}
\]

where $s(\mathbf{x},r)$ is the median absolute deviation of the boundary samples and $\overline{b}$ is the median. The final boundary response is computed as:

\[
R^{+}(\mathbf{x},r,\sigma)=R_{0}^{+}(\mathbf{x},r)S(\mathbf{x},r)^{\frac{3}{2}}
\]

\noindent To suppress responses at the border of vessels, the gradient magnitude
at the center of the vessel is combined with the offset medialness $R^{+}$ from above:

\[
R(\mathbf{x},r,\sigma)=\max\left\lbrace R^{+}(\mathbf{x},r)-\sigma|\nabla I^{\sigma}(\mathbf{x})|,0\right\rbrace
\]

\noindent The final vesselness response 
\[
R_{multi}(\mathbf{x})=\max_{\sigma,r}\left\lbrace R(\mathbf{x},r,\sigma)\right\rbrace 
\]
 is the maximum
response from all different scales $\sigma$ and radii $r$.
We found 4 scales and radii $r$ varying from 1 to 2 pixels, with an increase of 0.3 pixels to have the best performance.

\subsection{Centerline extraction}

In a non-maximum suppression step inspired by~\cite{Bauer2010172}, at each position $\mathbf{x}$ with a medialness $R(\mathbf{x})>th_{min}$, we sample 8 points
on a plane perpendicular to the estimated vessel direction. If the
medialness on any of those 8 points is larger than at the
current position $\mathbf{x}$, the VE response at $\mathbf{x}$ is
set to zero. This leads to disconnected vessel
centerline fragments due to branching points, where the tubularity 
assumption fails. Next, small centerline fragments (less than 5 $N_{26}$-connected
voxels) are removed, and all maxima lying on the airway border are
cleared. To reconnect the centerline fragments, we apply a Dijkstra-like
shortest path algorithm. In each lung separately, we connect all
centerline candidate points to the center of the heart. As a cost
function, we combine the medialness with the gradient magnitudes
of the CT image. The separate processing of right and left lung ensures that wrong connections through the mediastinum are avoided. The connected trees of the right and left lung form the final vessel tree.

\subsection{Vessel segmentation}

At each centerline voxel, 48 points lying on a sphere are sampled
from the CT image and the grey values are summed up. This is
done for spheres with different diameters. As soon as the normalized
sum of grey values drops under 0.6 (empirically found), the radius
of the vessel has been found. This gives a more accurate radius estimation
compared to the filter radii directly from the VE filter response. This coarse segmentation is then used as an input for a total variation based segmentation \cite{Reinbacher2010}, which gives the final segmentation.

\section{Experimental results\label{sec:Experimental-Results}}

For testing our algorithm we used the publicly available VESSEL12
challenge dataset as well as 24 datasets from patients who underwent contrast enhanced  CT as a part of a clinical PH study at the Ludwig Boltzmann Institute for Lung Vascular Research, Graz. The median size of the CT volumes is $512\times512\times426$ pixel. Our hardware consists of an Intel
CORE I7-2600K 3.40GHZ with 16 GB RAM and a Geforce 580 GTX with 3
GB RAM. 

All of our filters are implemented in CUDA, which is a parallel programming
model from NVIDIA, that generates hardware accelerated instructions for NVIDIA graphics processing units (GPUs). Using CUDA we can significantly improve the runtime of our 3D image processing algorithms, due to the high degree of parallelization. Therefore, we can directly work on the full resolution CT data. The runtime of the CUDA implementation of the whole algorithm pipeline ranges from 5 to 10 minutes, whereas a CPU-only implementation needs more than an hour. 
Limited memory on our GPU restricts the possible size of datasets in the different stages of our pipeline, so we decompose large (i.e. nearly $512^3$) CT images into overlapping sub-volumes, which are processed sequentially, with each sub-volume benefiting from the CUDA based parallelization.

\subsection{Airway and lung segmentation}

On 24 patients we were able to extract the airway tree to at least four generations, which is sufficient for our further processing. Due to GPU based parallelization, one airway segmentation takes on average 69 seconds. Figure~\ref{fig:Results-of-theairway}a show a representative result.

\begin{figure}[h!]
	\centering
	
	\subfigure[]{
			\includegraphics[height=3.5cm]{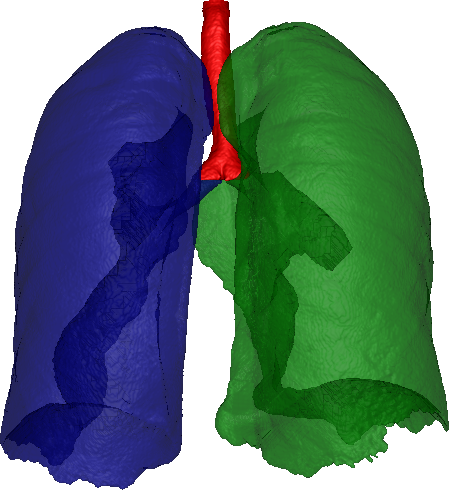}
		}
		\hfill
	\subfigure[]{
		\includegraphics[height=3.5cm]{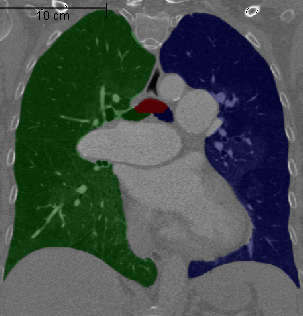}
		}
				\hfill
	\subfigure[]{
			\raisebox{5mm}{\includegraphics[width=0.3\textwidth]{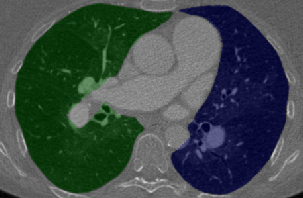}}
		}	
		
	\caption{(a) 3D rendering of an example lung segmentation, (b) coronal and (c) axial view; green=left lung, blue=right lung, red=trachea}
	\label{fig:Results-of-LungSeg}
\end{figure}

The lung segmentations from the 24 patients included all lung tissue and pulmonary vessels, which has been verified through visual inspection (Figure~\ref{fig:Results-of-LungSeg}). Processing time for the lung segmentation was on average 132 seconds.

\subsection{Phantom data set}
\begin{figure}[htb]
	\centering
	\subfigure[]{
				\raisebox{3mm}{\includegraphics[width=0.35\textwidth]{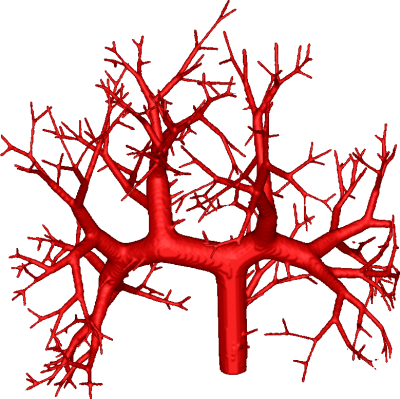}}
			}
	\hspace{1cm}
	\subfigure[]{
				\includegraphics[width=0.5\textwidth]{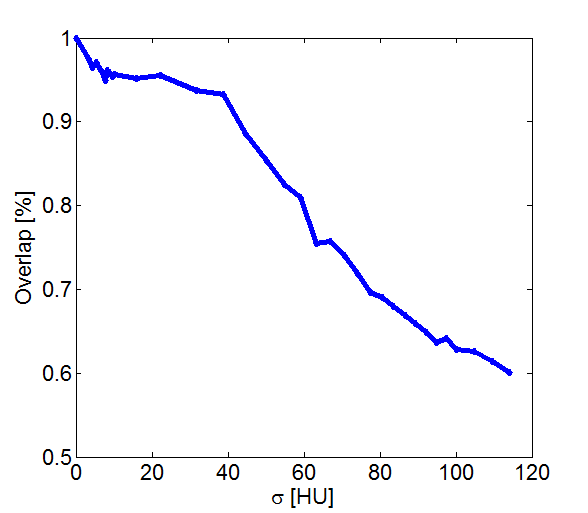}
			}

	\caption{(a) 3D rendering of liver vessel phantom, (b) Jaccard index over variance of Gaussian noise}
	\label{fig:Results-of-Phantom}
\end{figure}

We used a phantom of a liver vessel tree, depicted in Figure~\ref{fig:Results-of-Phantom}a, to check the performance of the algorithm and validate its robustness against noise. We successively added Gaussian
noise with increasing variance to the phantom data, and calculated the
Jaccard index of the ground-truth segmentation with the obtained segmentation. The curve in Figure~\ref{fig:Results-of-Phantom}b shows how the Jaccard index changes if Gaussian noise with increasing variance is added to the phantom dataset. As long as the variance of the noise is below $40$ Hounsfield Units (HU), the performance lies above 93\% segmentation overlap.

\subsection{VESSEL12 challenge}

We applied our algorithm to 20 datasets made available through the VESSEL12 challenge\footnote{http://vessel12.grand-challenge.org/}. Table ~\ref{tab:Vessel12Table} shows the results. The algorithm performs very well in terms of specificity, however improvement in the sensitivity is still necessary. This is because the algorithm is optimized for contrast enhanced CT images and for finding vessels even in noisy datasets. Thus vessels smaller than 2mm in diameter are misclassified as noise and are not included in the segmentation. All results from the other participating groups can be found on the official VESSEL12 challenge website.

\begin{table}[htb]
\centering
\caption{VESSEL12 challenge results; our team in comparison with the currently best performing team (LKEB China)}
\vspace{0.1cm}
\includegraphics[width=1\columnwidth]{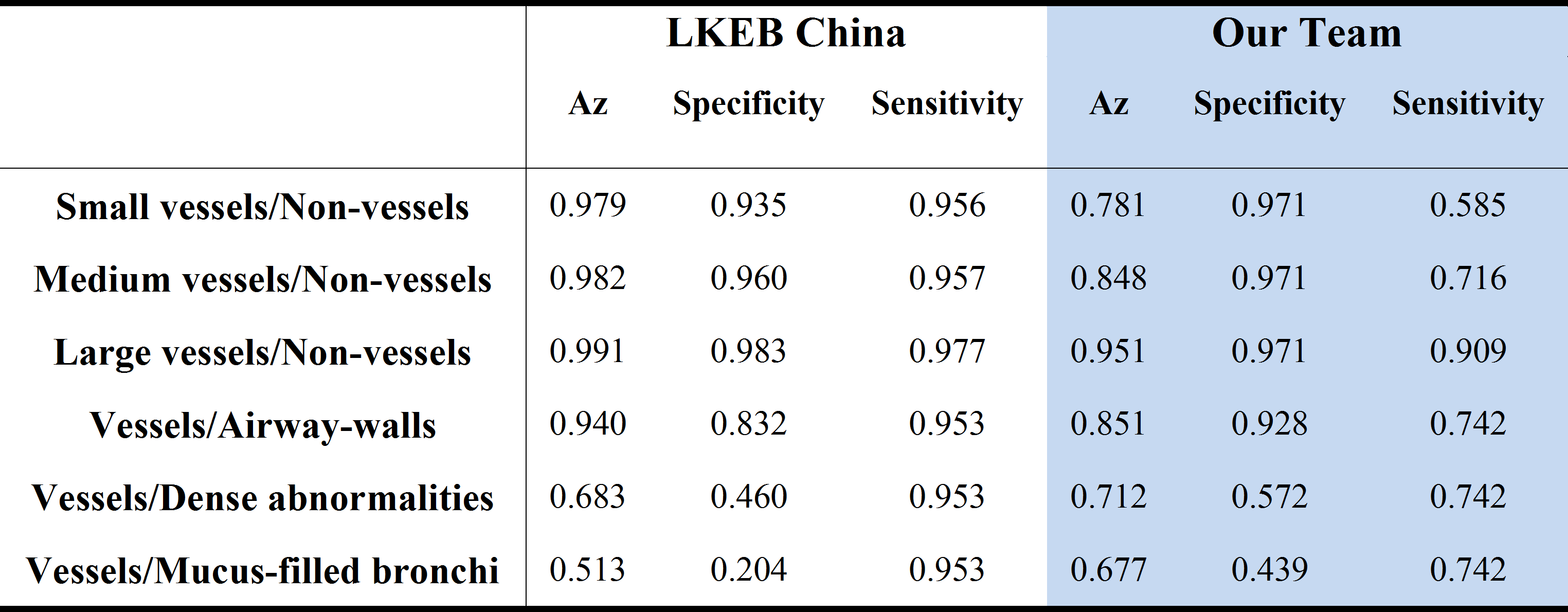}

\footnotesize{ Az: Area under the Receiver operating characteristic curve; Specificity: number of true negatives / (number of true negatives + number of false positives); Sensitivity: number of true positives / (number of true positives + number of false negatives)}

\label{tab:Vessel12Table}
\end{table}

\subsection{Contrast enhanced CT images from our clinical PH study}

Our clinical application is the detection of pulmonary hypertension
(PH), which is a chronic disorder of the pulmonary circulation, marked
by an elevated vascular resistance and elevated mean pulmonary artery
pressure (mPAP). Our hypothesis is, that the pulmonary vascular tree
shows quantifiable differences between patients with and without PH. One quantifiable property of the vessels is their
tortuosity, which is a readout of twistedness~\cite{Bullitt03measuringtortuosity}.
The most common metric of vascular tortuosity is the distance
metric (DM), which provides a ratio of the actual vessel
length to the Euclidean distance between its endpoints~\cite{Bullitt03measuringtortuosity}.
To determine the tortuosity, the lung vessel centerlines and branching points are extracted. The DM is calculated and compared
with the patient's clinical parameters. 

\begin{figure}[htb]
	\centering
	\subfigure[]{
					\includegraphics[height=3.5cm]{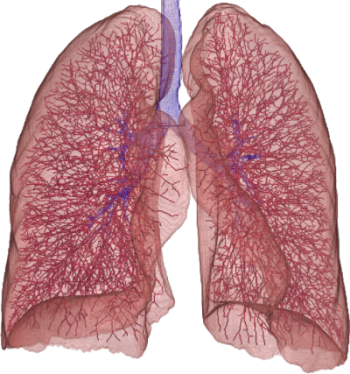}
				}
	\hspace{1cm}				
	\subfigure[]{
					\includegraphics[height=3.5cm]{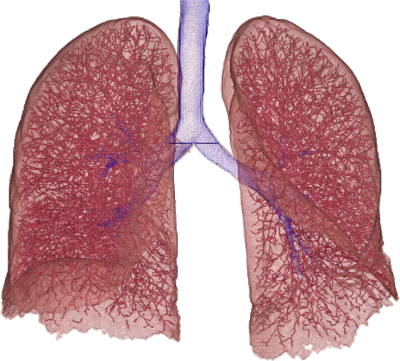}
				}
	
	\caption{Representative results of vessel segmentations showing a patient with (a) and one without PH (b). No visual differences between the vessel structure can be seen.}
	\label{fig:Results-of-PHStudyResults}
\end{figure}

We found a correlation between DM and mPAP of $\rho=0.60$ (Spearman correlation coefficient, p<0.01). There was a
significant difference between the DM of patients with and without PH (Table \ref{tab:Values-of-distance}, p<0.05), thus enabling to discriminate the two groups on our dataset of 24 patients. Two representative vessel segmentation results of the PH study datasets are shown in Figure~\ref{fig:Results-of-PHStudyResults}.

\begin{table}[htb]
\centering
\caption{Distance metric}
\vspace{0.1cm}
\includegraphics[width=1\columnwidth]{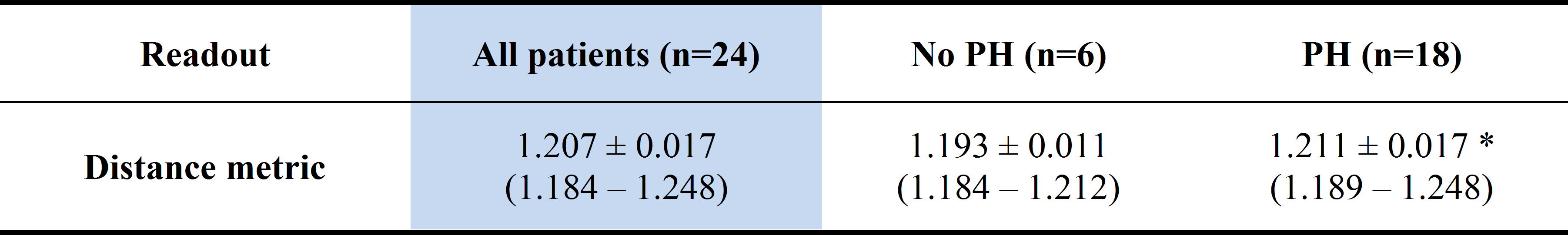}

\footnotesize{Data is presented as mean$\pm$standard deviation (range). The significance was tested with Students t-test; * p < 0.05 as compared to No PH group}

\label{tab:Values-of-distance}
\end{table}

\section{Conclusion}

We have presented a segmentation approach for vascular structures from contrast-enhanced CT images using a multi-scale vessel enhancement filter and using 
information from a lung- and airway-segmentation. We achieved very good segmentation results on our 24 patients from a clinical PH study. 
We also tested the algorithm on non-contrast-enhanced data from the VESSEL12 challenge, where we occupy a midfield position among all participating teams. We see room for improvement in the case of small vessels. Reasons for this performance of the algorithm are the optimization for the contrast-enhanced setup and the use of isotropic CT scans, which is not the case in the VESSEL12 datasets. 
Due to a parallelized CUDA implementation, our whole vessel tree segmentation and centerline extraction shows a run-time of at most 10 minutes for large CT datasets, without the need for computing on reduced resolutions, thus enabling the potential use in daily clinical routine.

As an important outcome of our work, we showed that tortuosity is correlated with mean pulmonary artery pressure, and our vessel segmentation algorithm can detect the presence of PH. One of the limitations of this study is the small number of patients, which allows only preliminary conclusions. A large scale prospective study to determine the true benefits and constraints of this method is currently in planning. Further, due to the radiation exposure one cannot test the repeatability
of the method. This would be necessary to determine its ability for
use in disease monitoring and follow-up examinations. 

\bibliography{refs}

\end{document}